%% file: paper.tex
\documentclass{article}

\usepackage{microtype}
\usepackage{subcaption}

\usepackage{graphicx}
\usepackage{booktabs} 
\usepackage{caption}

\usepackage{caption}
\usepackage{subcaption}
\usepackage{multirow}
\usepackage[table]{xcolor}
\usepackage{graphicx}
\usepackage{algorithm}
\usepackage{array}
\usepackage{fixltx2e}
\usepackage[table]{xcolor}
\usepackage{stfloats}
\usepackage{url}

\usepackage{times}
\usepackage{color}
\usepackage{verbatim}
\usepackage{subfiles}
\usepackage{lipsum}
\usepackage{xr}
\usepackage{hyperref}
\usepackage{zref-xr}
\usepackage{blindtext}
\usepackage{tabularx}
\usepackage{etoolbox}
\usepackage{setspace}
\usepackage{export}
\usepackage{nameref}
\usepackage{inputenc}
\usepackage{mathtools}
\usepackage{esvect}
\usepackage{xspace}	
\usepackage{soul}
\usepackage[export]{adjustbox}
\usepackage{amsmath, amsthm, amssymb}
\usepackage{comment}
\usepackage[compact,small]{titlesec}

\usepackage{titlesec}

\usepackage{hyperref}

\usepackage[accepted]{mlsys2021}

\titlespacing\section{0pt}{2pt plus 0pt minus 2pt}{0pt plus 2pt minus 2pt}
\titlespacing\subsection{0pt}{2pt plus 0pt minus 2pt}{0pt plus 2pt minus 2pt}
\titlespacing\subsubsection{0pt}{2pt plus 0pt minus 2pt}{0pt plus 2pt minus 2pt}

\mlsystitlerunning{TensorFlow Lite Micro}

\begin{document}

\twocolumn[
\mlsystitle{TensorFlow Lite Micro: \\ Embedded Machine Learning on TinyML Systems}



\mlsyssetsymbol{equal}{*}

\begin{mlsysauthorlist}
\mlsysauthor{Robert David}{goo}
\mlsysauthor{Jared Duke}{goo}
\mlsysauthor{Advait Jain}{goo}
\mlsysauthor{Vijay Janapa Reddi}{goo,hu}\\
\mlsysauthor{Nat Jeffries}{goo}
\mlsysauthor{Jian Li}{goo}
\mlsysauthor{Nick Kreeger}{goo}
\mlsysauthor{Ian Nappier}{goo}
\mlsysauthor{Meghna Natraj}{goo}\\
\mlsysauthor{Shlomi Regev}{goo}
\mlsysauthor{Rocky Rhodes}{goo}
\mlsysauthor{Tiezhen Wang}{goo}
\mlsysauthor{Pete Warden}{goo}
\end{mlsysauthorlist}

\mlsysaffiliation{goo}{Google}
\mlsysaffiliation{hu}{Harvard University}

\mlsyscorrespondingauthor{Pete Warden}{petewarden@google.com}
\mlsyscorrespondingauthor{Vijay Janapa Reddi}{vj@eecs.harvard.edu}

\mlsyskeywords{Machine Learning, tinyML, embedded machine learning, deep learning}

\vskip 0.2in

\begin{abstract}

\input{tex/abstract}
\end{abstract}
]



\printAffiliationsAndNotice{ }  

\input{tex/introduction}
\input{tex/technical_challenges}
\input{tex/design_principles}
\input{tex/implementation}
\input{tex/system_evaluation}
\input{tex/related_work}

\input{tex/conclusion}

\newpage
\input{tex/acknowledgements}

\bibliographystyle{mlsys2021}
\bibliography{references}



\end{document}

%% file: tex/abstract.tex
TensorFlow Lite Micro (TFLM) is an open-source ML inference framework for running deep-learning models on embedded systems. TFLM tackles the efficiency requirements imposed by embedded-system resource constraints and the fragmentation challenges that make cross-platform interoperability nearly impossible. The framework adopts a unique interpreter-based approach that provides flexibility while overcoming these unique challenges. In this paper, we explain the design decisions behind TFLM and describe its implementation. We present an evaluation of TFLM to demonstrate its low resource requirements and minimal run-time performance overheads.

%% file: tex/introduction.tex
\section{Introduction} \label{sec:introduction}

Tiny machine learning (TinyML) is a burgeoning field at the intersection of embedded systems and machine learning. The world has over 250 billion microcontrollers~\cite{icinsights}, with strong growth projected over coming years.
As such, a new range of embedded applications are emerging for neural networks.
Because these models are extremely small (few hundred KBs), running on microcontrollers or DSP-based embedded subsystems, they can operate continuously with minimal impact on device battery life. 

The most well-known and widely deployed example of this new TinyML technology is keyword spotting, also called hotword or wakeword detection \cite{chen2014small,gruenstein2017cascade,zhang2017hello}. Amazon, Apple, Google, and others use tiny neural networks on billions of devices to run always-on inferences for keyword detection---and this is far from the only TinyML application. Low-latency analysis and modeling of sensor signals from microphones, low-power image sensors, accelerometers, gyros, PPG optical sensors, and other devices enable consumer and industrial applications, including predictive maintenance \cite{pcoe,susto2014machine}, acoustic-anomaly detection \cite{Koizumi_WASPAA2019_01}, visual object detection \cite{chowdhery2019visual}, and human-activity recognition \cite{chavarriaga2013opportunity,zhang2012usc}.

Unlocking machine learning's potential in embedded devices requires overcoming two crucial challenges. First and foremost, embedded systems have no unified TinyML framework. When engineers have deployed neural networks to such systems, they have built one-off frameworks that require manual optimization for each hardware platform. Such custom frameworks have tended to be narrowly focused, lacking features to support multiple applications and lacking portability across a wide range of hardware. The developer experience has therefore been painful, requiring hand optimization of models to run on a specific device. And altering these models to run on another device necessitated manual porting and repeated optimization effort. An important second-order effect of this situation is that the slow pace and high cost of training and deploying models to embedded hardware prevents developers from easily justifying the investment required to build new features.

Another challenge limiting TinyML is that hardware vendors have related but separate needs. Without a generic TinyML framework, {evaluating hardware performance in a neutral, vendor-agnostic manner has been difficult}. Frameworks are tied to specific devices, and it is hard to determine the source of improvements because they can come from hardware, software, or the complete vertically integrated solution.

The lack of a proper framework has been a barrier to accelerating TinyML adoption and application in products. Beyond deploying a model to an embedded target, the framework must also have a means of training a model on a higher-compute platform. TinyML must exploit a broad ecosystem of tools for ML, as well for orchestrating and debugging models, which are beneficial for production devices. 

Prior efforts have attempted to bridge this gap. We can distill the major issues facing the frameworks into the following:
\begin{itemize}
  \setlength\itemsep{-0.085em}

    \item Inability to easily and portably deploy models across multiple embedded hardware architectures
    \item Lack of optimizations that take advantage of the underlying hardware without requiring framework developers to make platform-specific efforts
    \item Lack of productivity tools that connect training pipelines to deployment platforms and tools
    \item Incomplete infrastructure for compression, quantization, model invocation, and execution
    \item Minimal support features for performance profiling, debugging, orchestration, and so on
    \item No benchmarks that allow vendors to quantify their chip's performance in a fair and reproducible manner
    \item Lack of testing in real-world applications.
\end{itemize}

To address these issues, we introduce TensorFlow Lite Micro (TFLM), which mitigates the slow pace and high cost of training and deploying models to embedded hardware by emphasizing portability and flexibility. TFLM makes it easy to get TinyML applications running across architectures, and it allows hardware vendors to incrementally optimize kernels for their devices. It gives vendors a neutral platform to prove their performance and offers these benefits:

\begin{itemize}
  \setlength\itemsep{-0.085em}
    \item Our interpreter-based approach is portable, flexible, and easily adapted to new applications and features
    \item We minimize the use of external dependencies and library requirements to be hardware agnostic
    \item We enable hardware vendors to provide platform-specific optimizations on a per-kernel basis without writing target-specific compilers
    \item We allow hardware vendors to easily integrate their kernel optimizations to ensure performance in production and comparative hardware benchmarking
    \item Our model-architecture framework is open to a wide machine-learning ecosystem and the TensorFlow Lite model conversion and optimization infrastructure 
    \item We provide benchmarks that are being adopted by industry-leading benchmark bodies like MLPerf
    \item Our framework supports popular, well-maintained Google applications that are in production.
\end{itemize}

This paper makes several contributions: First, we clearly lay out the challenges to developing a machine-learning framework for embedded devices that supports the fragmented embedded ecosystem. Second, we provide design and implementation details for a system specifically created to cope with these challenges. And third, we demonstrate that an interpreter-based approach, which is traditionally viewed as a low-performance alternative to compilation, is in fact highly suitable for the embedded domain---specifically, for machine learning. Because machine-learning performance is largely dictated by linear-algebra computations, the interpreter design imposes minimal run-time overhead. 


%% file: tex/technical_challenges.tex
\section{Technical Challenges} \label{sec:technical_challenges}

Many issues make developing an ML framework for embedded systems particularly difficult, as discussed here. 


\subsection{Missing Features} \label{subsec:missing_features}

Embedded platforms are defined by their tight limitations. Therefore, many advances from the past few decades that have made software development faster and easier are unavailable to these platforms because the resource tradeoffs are too expensive. Examples include dynamic memory management, virtual memory, an operating system, a standard instruction set, a file system, floating-point hardware, and other tools that seem fundamental to modern programmers~\cite{kumar2017resource}. Though some platforms provide a subset of these features, a framework targeting widespread adoption in this market must avoid relying on them.

\subsection{Fragmented Market and Ecosystem} \label{subsec:fragmented_market}


Many embedded-system uses only require fixed software developed alongside the hardware, usually by an affiliated team. The lack of applications capable of running on the platform is therefore much less important than it is for general-purpose computing. Moreover, backward instruction-set-architecture (ISA) compatibility with older software matters less than in mainstream systems because everything that runs on an embedded system is probably compiled from source code anyway. Thus, embedded hardware can aggressively diversify to meet power requirements, whereas even the latest x86 processor can still run instructions that are nearly three decades old \cite{intel2013intel}.

These differences mean the pressure to converge on one or two dominant platforms or ISAs is much weaker in the embedded space, leading to fragmentation. Many ISAs have thriving ecosystems, and the benefits they bring to particular applications outweigh developers' cost of switching. Companies even allow developers to add their own ISA extensions~\cite{waterman2019risc,arm-custom}.

Matching the wide variety of embedded architectures are the numerous tool chains and integrated development environments (IDEs) that support them. Many of these systems are only available through a commercial license with the hardware manufacturer, and in cases where a customer has requested specialized instructions, they may be inaccessible to everyone. These arrangements have no open-source ecosystem, leading to device fragmentation that prevents a lone development team from producing software that runs well on many different embedded platforms.

\subsection{Resource Constraints}

People who build embedded devices do so because a general-purpose computing platform exceeds their design limits. The biggest drivers are cost, with a microcontroller typically selling for less than a few dollars \cite{icinsights}; power consumption, as embedded devices may require just a few milliwatts of power, whereas mobile and desktop CPUs require watts; and form factor, since capable microcontrollers are smaller than a grain of rice \cite{wu20180}.

To meet their needs, hardware designers trade off capabilities. A common characteristic of an embedded system is its low memory capacity. At one end of the spectrum, a big embedded system has a few megabytes of flash ROM and at most a megabyte of SRAM. At the other end, a small embedded system  has just a few hundred kilobytes or fewer, often split between ROM and RAM \cite{zhang2017hello}. 

These constraints mean both working memory and permanent storage are much smaller than most software written for general-purpose platforms would assume. In particular, the size of the compiled code in storage requires minimization. 


Most software written for general-purpose platforms contains code that often goes uncalled on a given device. Choosing the code path at run time is a better use of engineering resources than shipping more-highly custom executables. Such run-time flexibility is hard to justify when code size is a concern and the potential uses are fewer. As a result, developers must break through the a library's abstraction if they want to make modifications to suit their target hardware.

\subsection{Ongoing Changes to Deep Learning}


Machine learning remains in its infancy despite its breakneck pace. Researchers are still experimenting with new operations and network architectures to glean better predictions from their models. Their success in improving results leads product designers to demand these enhanced models. 

Because new mathematical operations---or other fundamental changes to neural-network calculations---often drive the model advances, adopting these models in software means porting the changes, too. Since research directions are hard to predict and advances are frequent, keeping a framework up to date and able to run the newest, best models requires a lot of work. Hence, for instance, while TensorFlow has more than 1,400 operations~\cite{tfops}, TensorFlow Lite, which is deployed on more than four billions edge devices worldwide, supports only about 130 operations. Not all operations are worth supporting, however.

%% file: tex/design_principles.tex
\section{Design Principles} \label{sec:design_principles}

To address the challenges, we developed a set of developer principles to guide the design of TFLM that we discuss here.

\subsection{Minimize Feature Scope for Portability}

We believe an embedded machine-learning (ML) framework should assume the model, input data, and output arrays are in memory, and it should only handle ML calculations based on those values. The design should exclude any other function, no matter how useful. In practice, this approach means the library should omit features such as loading models from a file system or accessing peripherals for inputs. 

This  principle is crucial as many embedded platforms are missing basic features, such as memory management and library support (Section~\ref{subsec:missing_features}), that mainstream platforms take for granted. Supporting the myriad possibilities would make porting the ML framework across devices unwieldy. 

Fortunately, ML models are functional, having clear inputs, outputs, and possibly some internal state but no external side effects. Running a model need not involve calls to peripherals or other operating-system functions. To remain efficient, we focus only on implementing those calculations. 

\subsection{Enable Vendor Contributions to Span Ecosystem}

All embedded devices can benefit from high-performance kernels optimized for a given microprocessor. But no one team can easily support such kernels for the entire embedded market because of the ecosystem's fragmentation (see Section~\ref{subsec:fragmented_market}). Worse, optimization approaches vary greatly depending on the target microprocessor architecture.

The companies with the strongest motivation to deliver maximum performance on a set of devices are the ones that design and sell the underlying embedded microprocessors. Although developers at these companies are highly experienced at optimizing traditional numerical algorithms (e.g., digital signal processing) for their hardware, they often lack deep-learning experience. Therefore, evaluating whether their optimization changes are detrimental or acceptable to model accuracy and overall performance is difficult.

To improve the development experience for hardware vendors and application developers, we make sure optimizing the core library operations is easy. One goal is to ensure substantial technical support (tests and benchmarks) for developer modifications and to encourage submission to a library repository (details are presented in Section~\ref{sec:implementation}).


\subsection{Reuse TensorFlow Tools for Scalability}



The TensorFlow training environment includes more than 1,400 operations, similar to other training frameworks~\cite{tfops}. Most inference frameworks, however, explicitly support only a subset of these operations, making exports difficult. An exporter takes a trained model (such as a TensorFlow model) and generates a TensorFlow Lite model file (.tflite); after conversion, the model file can be deployed to a client device (e.g., a mobile or embedded system) and run locally using the TensorFlow Lite interpreter. 

Exporters receive a constant stream of new operations, most defined only by their implementation code. Because the operands lack clean semantic definitions beyond their implementations and unit tests, supporting these operations is difficult. Attempting to do so is like working with the elaborate CISC ISA without access to the ISA manual.

Manually converting/exporting one or two models to a new representation is easy.
Users will want to convert a large space of potential models, however, and the task of understanding and changing model architectures to accommodate a framework's requirements is difficult. 
Often, only after users have built and trained a model do they discover whether all of its operations are compatible with the target inference framework. Worse, many users employ high-level APIs, such as Keras \cite{chollet2015keras}, which may hide low-level operations, complicating the task of removing depencence on operations. Also, researchers and product developers often split responsibilities, with the former creating models and the latter deploying them. Since product developers are the ones who discover the export errors, they may lack the expertise or permission to retrain the model. 

Model operators have no governing principles or a unified set of rules. Even if an inference framework supports an operation, particular data types may not, or the operation may exclude certain parameter ranges or may only serve in conjunction with other operations. This situation creates a barrier to providing error messages that guide developers.

Resource constraints also add many requirements to an exporter. Most training frameworks focus on floating-point calculations, since they are the most flexible numerical representation and are well optimized for desktop CPUs and GPUs. Fitting into small memories, however, makes eight-bit and other quantized representations valuable for embedded deployment. Some techniques can convert a model trained in floating point to a quantized representation \cite{krishnamoorthi2018quantizing}, but they all increase exporter complexity. Some also require support during the training process, necessitating changes to the creation framework as well. Other optimizations are also expected during export, such as folding constant expressions into fixed values---even in complex cases like batch normalization~\cite{zhang2017hello}---and removing dropout and similar operations that are only useful during training \cite{srivastava2014dropout}.

\begin{figure}[t]
    \centering
    \includegraphics[trim=0 0 0 0, clip, width=\linewidth]{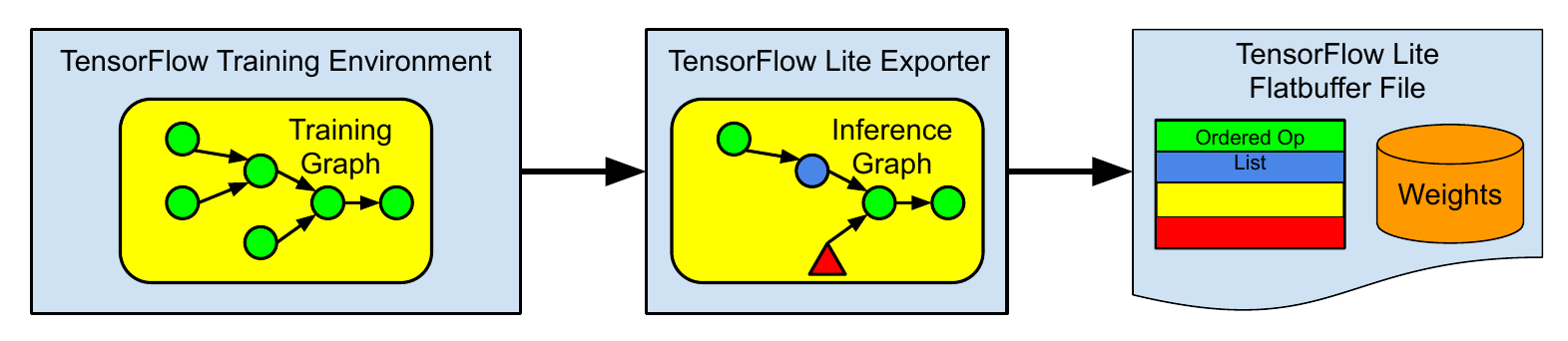}
    \vspace{-1em}
    \caption{Model-export workflow.}
    \label{fig:model_export_workflow}
\end{figure}

Because writing a robust model converter takes a tremendous amount of engineering work, we built atop the existing TensorFlow Lite tool chain. As Figure~\ref{fig:model_export_workflow} shows, we use the TensorFlow Lite toolchain to ease conversion and optimization and the converter outputs a FlatBuffer file used by TFLM to load the inference models. We exploited this strong integration with the TensorFlow training environment and extended it for rapidly supporting deeply embedded machine-learning systems. For example, we reuse the TensorFlow Lite reference kernels, thus giving users a harmonized environment for model development and execution. 


\subsection{Build System for Heterogeneous Support} 
\label{subsec:build_source_files}

A crucial feature is a flexible build environment. The build system must support the highly heterogeneous ecosystem and avoid falling captive to any one platform. Otherwise, developers would avoid adopting it due to the lack of portability and so would the hardware platform vendors.

In desktop and mobile systems, frameworks commonly provide precompiled libraries and other binaries as the main software-delivery method. This approach is impractical in embedded platforms because they encompass too many different devices, operating systems, and tool-chain combinations to allow a balancing of modularity, size, and other constraints. 
Additionally, embedded system developers must often make code changes to meet such constraints.

We prioritize code that is easy to build using various IDEs and tool chains. This approach means we avoid techniques that rely on build-system features that do not genearlize across platforms. Examples of such features include setting custom include paths, compiling tools for the host processor, using custom binaries or shell scripts to produce code, and defining preprocessor macros on the command line.

Our principle is that we should be able to create source files and headers for a given platform, and users should then be able to drag and drop those files into their IDE or tool chain and compile them without any changes. We call it the ``Bag of Files'' principle. Anything more complex would prevent adoption by many platforms and developers.

%% file: tex/implementation.tex
\section{Implementation} \label{sec:implementation}

We discuss our implementation decisions and tradeoffs we make as we describe specific modules in detail. 


\subsection{System Overview}

The first step in developing a TFLM application is to create a live neural-network-model object in memory. The application developer produces an ``operator resolver'' object through the client API. The ``OpResolver'' API controls which operators link to the final binary, minimizing file size.

The second step is to supply a contiguous memory ``arena'' that holds intermediate results and other variables the interpreter needs. Doing so is necessary because we assume dynamic memory allocation is unavailable.

The third step is to create an interpreter instance (Section~\ref{subsec:micro_interpreter}), supplying it with the model, operator resolver, and arena as arguments. The interpreter allocates all required memory from the arena during the initialization phase. We avoid any allocations afterward to ensure heap fragmentation avoids causing errors for long-running applications. Operator implementations may allocate memory for use during the evaluation, so the operator preparation functions are called during this phase, allowing their memory needs to be communicated to the interpreter. The application-supplied OpResolver maps the operator types listed in the serialized model to the implementation functions. A C API call handles all communication between the interpreter and operators to ensure operator implementations are modular and independent of the interpreter’s details. This approach eases replacement of operator implementations with optimized versions, and it also encourages reuse of other systems' operator libraries (e.g., as part of a code-generation project).

The fourth step is execution. The application retrieves pointers to the memory regions that represent the model inputs and populates them with values (often derived from sensors or other user-supplied data). Once the inputs are available, the application invokes the interpreter to perform the model calculations. This process involves iterating through the topologically sorted operations, using offsets calculated during memory planning to locate the inputs and outputs, and calling the evaluation function for each operation. 

Finally, after it evaluates all the operations, the interpreter returns control to the application. Invocation is a simple blocking call. Most MCUs are single-threaded and they use interrupts for urgent tasks so it is acceptable. But an application can still perform one from a thread, and platform-specific operators can still split their work across processors. Once invocation finishes, the application can query the interpreter to determine the location of the arrays containing the model-calculation outputs and then use those outputs.

The framework omits any threading or multitasking support, since any such features would require less-portable code and operating-system dependencies. However, we support multitenancy. The framework can run multiple models as long as they do not need to run concurrently with one another. 

\subsection{TFLM Interpreter} \label{subsec:micro_interpreter}

TFLM is an interpreter-based machine-learning inference framework. The interpreter loads a data structure that clearly defines a machine learning model. Although the execution code is static, the interpreter handles the model data at run time, and this data controls which operators to execute and where to draw the model parameters from.

\begin{figure}[t]
    \centering
    \includegraphics[trim=0 0 0 0, clip, width=0.75\linewidth]{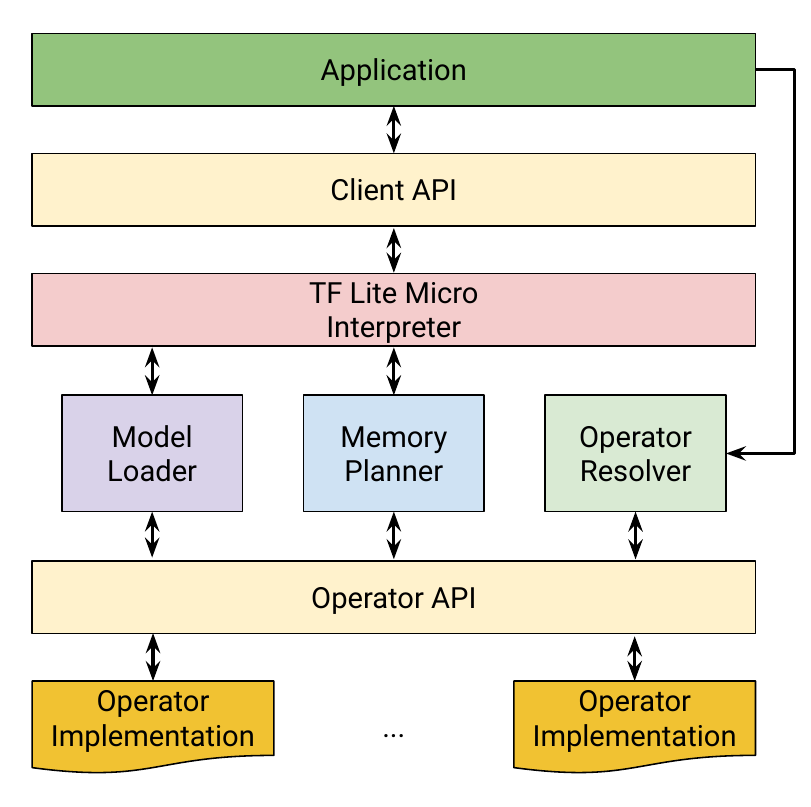}
    \caption{Implementation-module overview.}
    \label{fig:impl_overview}
\end{figure}



We chose an interpreter on the basis of our experience deploying production models on embedded hardware. We see a need to easily update models in the field---a task that may be infeasible using code generation. Using an interpreter, however, sharing code across multiple models and applications is easier, as is maintaining the code, since it allows updates without re-exporting the model. Moreover, unlike traditional interpreters with lots of branching overhead relative to a function call, ML model interpretation benefits from long-running kernel complexity. Each kernel runtime is large and amortizes the interpreter overhead (Section~\ref{sec:system_evaluation}).

The alternative to an interpreter-based inference engine is to generate native code from a model during export using C or C++, baking operator function calls into fixed machine code. It can increase performance at the expense of portability, since the code would need recompilation for each target. Code generation intersperses settings such as model architecture, weights, and layer dimensions in the binary, which means replacing the entire executable to modify a model. In contrast, an interpreted approach keeps all this information in a separate memory file/area, allowing model updates to replace a single file or contiguous memory area.


We incorporate some important code-generation features in our approach. For example, because our library is buildable from source files alone (Section~\ref{subsec:build_source_files}), we achieve much of the compilation simplicity of generated code. 

\subsection{Model Loading} \label{subsec:model_loading}

As mentioned, the interpreter loads a data structure that clearly defines a model. For this work, we used the TensorFlow Lite portable data schema~\cite{tfliteguide}. Reusing the export tools from TensorFlow Lite enabled us to import a wide variety of models at little engineering cost.  

\subsubsection{Model Serialization} TensorFlow Lite for smartphones and other mobile devices employs the FlatBuffer serialization format to hold models \cite{tfliteflatbuffers}. The binary footprint of the accessor code is typically less than two kilobytes. It is a header-only library, making compilation easy, and it is memory efficient because the serialization protocol does not require unpacking to another representation.
The downside to this format is that its C++ header requires the platform compiler to support the C++11 specification. 

We had to work with several vendors to upgrade their tool chains to handle this version, but since we had implicitly chosen modern C++ by basing our framework on TensorFlow Lite, it has been a minor obstacle. Another challenge of this format was that most of the target embedded  devices lacked file systems, but because it uses a memory-mapped representation, files are easy to convert into C source files containing data arrays. These files are compilable into the binary, to which the application can easily refer.

\subsubsection{Model Representation} We also copied the TensorFlow Lite representation, the stored schema of data and values that represent the model. This schema was designed for mobile platforms with storage efficiency and fast access in mind, so it has many features that eased development for embedded platforms. For example, operations reside in a topologically sorted list rather than a directed-acyclic graph. Performing calculations is as simple as looping through the operation list in order,
whereas a full graph representation would require preprocessing to satisfy the operations' input dependencies.


The drawback of this representation is that it was designed to be portable from system to system, so it requires run-time processing to yield the information that inferencing requires. For example, it abstracts operator parameters from the arguments, which later pass to the functions that implement those operations. Thus, each operation requires a few code lines executed at run time to convert from the serialized representation to the structure in the underlying implementation. The code overhead is small, but it reduces the readability and compactness of the operator implementations. 

Memory planning is a related issue. On mobile devices, TensorFlow Lite supports variable-size inputs, so all dependent operations may also vary in size. Planning the optimal layout of intermediate buffers for the calculations must take place at run time when all buffer dimensions are known. 


\subsection{Memory Management}


We are unable to assume the operating system can dynamically allocate memory. So the framework allocates and manages memory from a provided memory arena. During model preparation, the interpreter determines the lifetime and size of all buffers necessary to run the model. These buffers include run-time tensors, persistent memory to store metadata, and scratch memory to temporarily hold values while the model runs (Section~\ref{subsec:scratchpad}). After accounting for all required buffers, the framework creates a memory plan that reuses nonpersistent buffers when possible while ensuring buffers are valid during their required lifetime (Section~\ref{subsec:memory_planning}).

\subsubsection{Persistent Memory and Scratchpads}
\label{subsec:scratchpad}

We require applications to supply a fixed-size memory arena when they create the interpreter and to keep the arena intact throughout the interpreter's lifetime. Allocations with the same lifetime can treat this arena as a stack. 
If an allocation takes up too much space, we raise an application-level error.

To prevent memory errors from interrupting a long-running program, we ensure that allocations only occur during the interpreter's initialization phase. No allocation (through our mechanisms) is possible during model invocation.

This simplistic approach works well for initial prototyping, but it wastes memory because many allocations could overlap with others in time. One example is data structures that are only necessary during initialization. Their values are irrelevant after initialization, but because their lifetime is the same as the interpreter's, they continue to take up arena space. A model's evaluation phase also requires variables that need not persist from one invocation to another.

\begin{figure}[t]
    \centering
    \includegraphics[trim=0 0 0 0, clip, width=\linewidth]{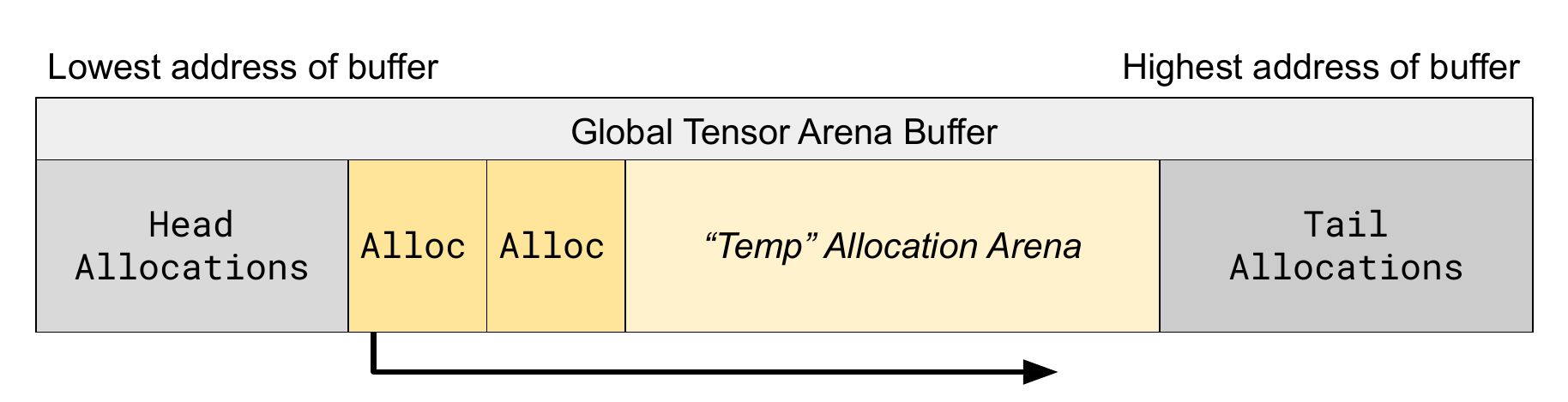}
    \caption{Two-stack allocation strategy.}
    \label{fig:two_stack_alloc}
\end{figure}

Hence, we modified the allocation scheme so that initialization- and evaluation-lifetime allocations reside in a separate stack relative to interpreter-lifetime objects. This feat uses a stack that increments from the lowest address for the function-lifetime objects (``Head'' in Figure~\ref{fig:two_stack_alloc}) and a stack that decrements from the arena's highest address for interpreter-lifetime allocations (``Tail'' in Figure~\ref{fig:two_stack_alloc}). When the two stack pointers cross, they indicate a lack of capacity.

The two-stack allocation strategy works well for both shared buffers and persistent buffers. But model preparation also holds allocation data that model inference no longer needs. Therefore, we used the space in between the two stacks as temporary allocations when a model is in memory planning. Any temporary data required during model inference resides in the persistent-stack allocation section.

Our approach reduces the arena size as the initialization allocations can be discarded after that function is done, and the memory is reusable for evaluation variables. This approach also enables advanced applications to reuse the arena's function-lifetime section in between evaluation calls.

\subsubsection{Memory Planner}
\label{subsec:memory_planning}
A more complex optimization opportunity involves the space required for intermediate calculations during model evaluation. An operator may write to one or more output buffers, and later operators may later read them as inputs. If the output is not exposed to the application as a model output, its contents need only remain until the last operation that needs them has finished. Its presence is also unnecessary until just before the operation that populates it executes. Memory reuse is possible by overlapping allocations that are unneeded during the same evaluation sections. 

\begin{figure}[t]
\centering
    \begin{subfigure}{0.7\columnwidth}
        \vskip 0pt
        \centering
        \includegraphics[valign=t, trim=0 0 0 0, clip, width=\linewidth]{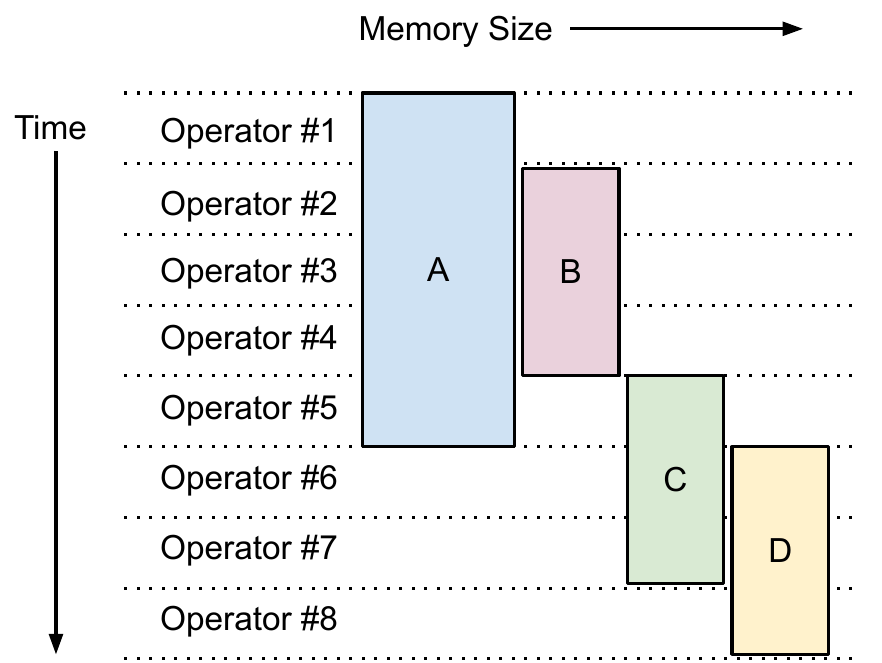}
        \caption{Naive}
        \label{fig:naive_inter_alloc}
    \end{subfigure}
    \hfill
    \begin{subfigure}{0.29\columnwidth}
        \vskip 0pt
        \centering
        \includegraphics[valign=t, trim=0 0 0 0, clip, width=\linewidth]{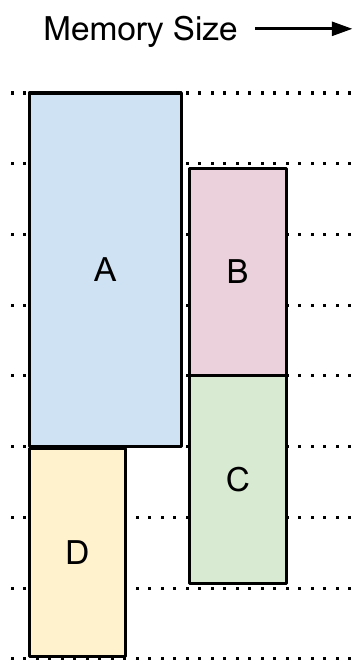}
        \caption{Bin packing}
        \label{fig:optimal_inter_alloc}
    \end{subfigure}
    \caption{Intermediate allocation strategies.}
\end{figure}

The memory allocations required over time can be visualized using rectangles (Figure~\ref{fig:naive_inter_alloc}), where one dimension is memory size and the other is the time during which each allocation must be preserved. The overall memory can be substantially reduced if some areas are reused or compacted together. Figure~\ref{fig:optimal_inter_alloc} shows a more optimal memory layout.

Memory compaction is an instance of bin packing \cite{martello1990knapsack}. Calculating the perfect allocation strategy for arbitrary models without exhaustively trying all possibilities is an unsolved problem, but a first-fit decreasing algorithm \cite{garey1972worst} usually provides reasonable solutions. 

In our case, this approach consists of gathering a list of all temporary allocations, including size and lifetime; sorting the list in descending order by size; and placing each allocation in the first sufficiently large gap, or at the end of the buffer if no such gap exists. We do not support dynamic shapes in the TFLM framework, so we must know at initialization all the information necessary to perform this algorithm. The ``Memory Planner'' (shown in Figure~\ref{fig:impl_overview}) encapsulates this process; it allows us to minimize the arena portion devoted to intermediate tensors. Doing so offers a substantial memory-use reduction for many models.

Memory planning at run time incurs more overhead during model preparation than a preplanned memory-allocation strategy. This cost, however, comes with the benefit of model generality. TFLM models simply list the operator and tensor requirements. At run time, we allocate and enable this capability for many model types.


Offline-planned tensor allocation is an alternative memory-planning feature of TFLM. It allows a more compact memory plan, gives memory-plan ownership and control to the end user, imposes less overhead on the MCU during initialization, and enables more-efficient power options by allowing different memory banks to store certain memory areas. We allow the user to create a memory layout on a host before run time. The memory layout is stored as model FlatBuffer metadata and contains an array of fixed-memory arena offsets for an arbitrary number of variable tensors. 

\subsection{Multitenancy}

Embedded-system constraints can force application-model developers to create several specialized models instead of one large monolithic model. Hence, supporting multiple models on the same embedded system may be necessary. 

If an application has multiple models that need not run simultaneously, it is possible to have two separate instances running in isolation from one another. However, this is inefficient because the temporary space cannot be reused. 

Instead, TFLM supports multitenancy with some memory-planner changes that are transparent to the developer. TFLM supports memory-arena reuse by enabling the multiple model interpreters to allocate memory from a single arena. 

We allow interpreter-lifetime areas to stack on each other in the arena and reuse the function-lifetime section for model evaluation. The reusable (nonpersistent) part is set to the largest requirement, based on all models allocating in the arena. The nonreusable (persistent) allocations grow for each model---allocations are model specific (Figure~\ref{fig:optimal_inter_alloc}).

\begin{figure*}[t]
\centering
    \begin{subfigure}{\columnwidth}
        \centering
        \includegraphics[trim = 0 0 0 0, clip, width=\linewidth]{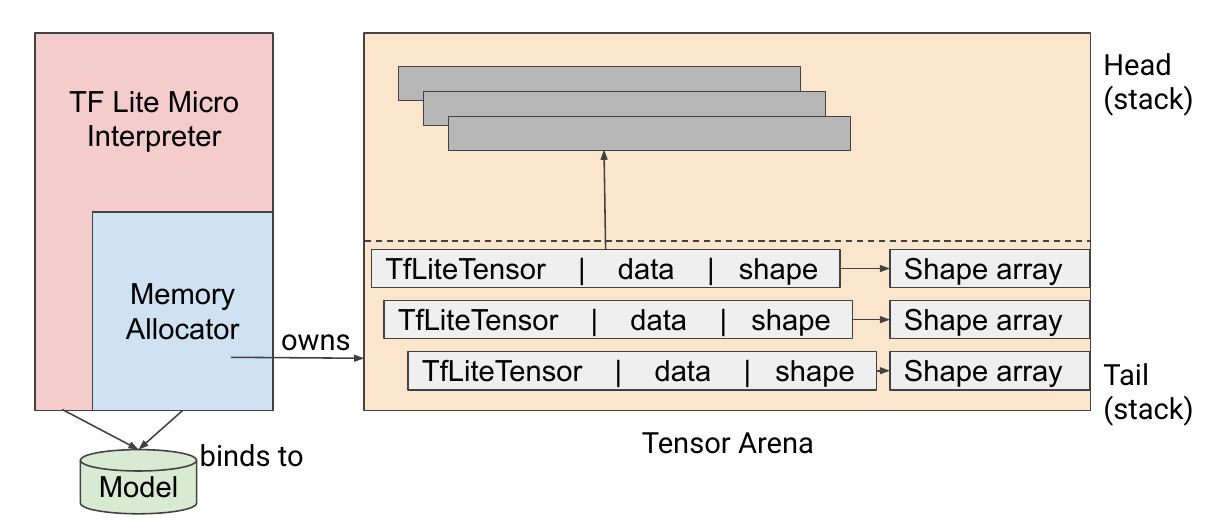}
        \caption{Single-model}
        \label{fig:single_model}
    \end{subfigure}
    \begin{subfigure}{\columnwidth}
        \centering
        \includegraphics[trim = 0 0 0 0, clip, width=\linewidth]{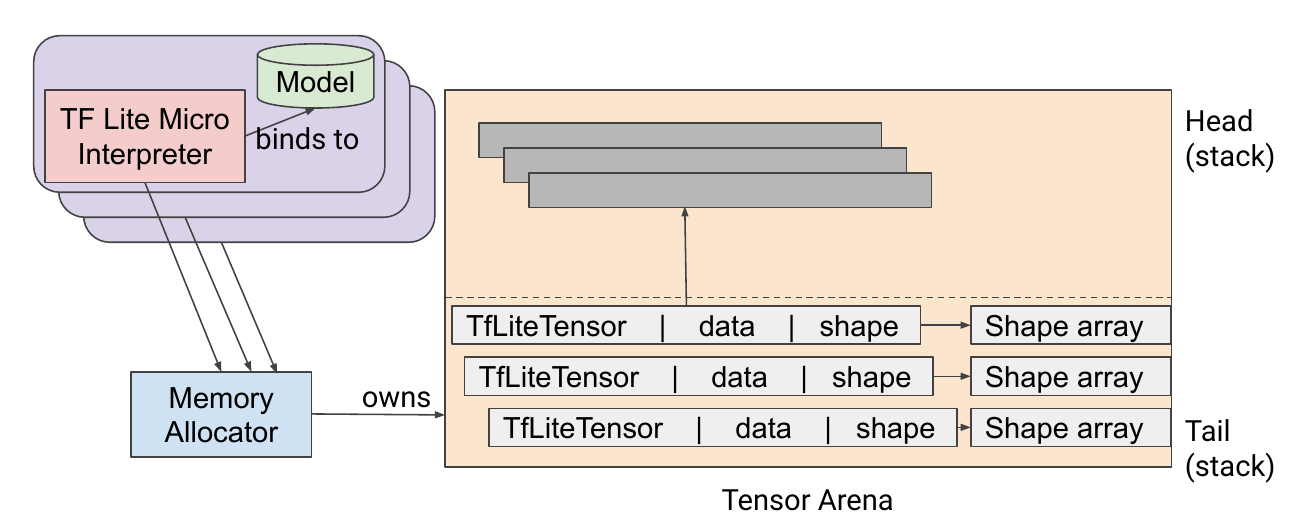}
        \caption{Multiple models.}
        \label{fig:multi_model}
    \end{subfigure}
    \caption{Memory-allocation strategy for a single model versus a multi-tenancy scenario. In TFLM, there is a one-to-one binding between a model, an interpreter and the memory allocations made for the model (which may come from a shared memory arena).}
\end{figure*}

\subsection{Multithreading}

TFLM is thread-safe as long as there is no state corresponding to the model that is kept outside the interpreter and the model's memory allocation within the arena.

The interpreter's only variables are kept in the arena, and each interpreter instance is uniquely bound to a specific model. Therefore, TFLM can safely support multiple interpreter instances running from different tasks or threads.

TFLM can also run safely on multiple MCU cores. Since the only variables used by the interpreter are kept in the arena, this works well in practice. The executable code is shared, but the arenas ensure there are no threading issues.



\subsection{Operator Support}

Operators are the calculation units in neural-network graphs. They represent a sizable amount of computation, typically requiring many thousands or even millions of individual arithmetic operations (e.g., multiplies or additions).
They are functional, with well-defined inputs, outputs, and state variables as well as no side effects beyond them.

Because the model execution's latency, power consumption, and code size tend to be dominated by the implementations of these operations, they are typically specialized for particular platforms to take advantage of hardware characteristics. In practice, we attracted library optimizations from hardware vendors such as Arm, Cadence, Ceva, and Synopsys.

Well-defined operator boundaries mean it is possible to define an API that communicates the inputs and outputs but hides implementation details behind an abstraction. Several chip vendors have provided a library of neural network kernels designed to deliver maximum neural-network performance when running on their processors. For example, Arm has provided optimized CMSIS-NN libraries divided into several functions, each covering a category: convolution, activation, fully connected layer, pooling, softmax, and optimized basic math. TFLM uses CMSIS-NN to deliver high performance as we demonstrate in Section~\ref{sec:system_evaluation}. 


\subsection{Platform Specialization}

TFLM gives developers flexibility to modify the library code. Because operator implementations (kernels) often consume the most time when executing models, they are prominent targets for platform-specific optimization. 

We wanted to make swapping in new implementations easy. To do so, we allow specialized versions of the C++ source code to override the default reference implementation. Each kernel has a reference implementation that is in a directory, but subfolders contain optimized versions for particular platforms (e.g., the Arm CMSIS-NN library). 

As we explain in Section~\ref{subsec:build_system}, the platform-specific source files replace the reference implementations during all build steps when targeting the named platform or library (e.g., using \texttt{TAGS="cmsis-nn"}). Each platform is given a unique tag. The tag is a command line argument to the build system that replaces the reference kernels during compilation. 
In a similar vein, library modifiers can swap or change the implementations incrementally with no changes to the build scripts and the overarching build system we put in place.

\subsection{Build System} \label{subsec:build_system}

To address the embedded market's fragmentation (Section~\ref{subsec:fragmented_market}), we needed our code to compile on many platforms. We therefore wrote the code to be highly portable, exhibiting few dependencies, but it was insufficient to give potential users a good experience on a particular device. 

Most embedded developers employ a platform-specific IDE or tool chain that abstracts many details of building subcomponents and presents libraries as interface modules. Simply giving developers a folder hierarchy containing source-code files would still leave them with multiple steps before they could build and compile that code into a usable library.

Therefore, we chose a single \texttt{makefile} based build system to determine which files the library required, then generated the project files for the associated tool chains. The makefile held the source-file list, and we stored the platform-specific project files as templates that the project-generation process filled in with the source-file information. That process may also perform other postprocessing to convert the source files to a format suitable for the target tool chain.

Our platform-agnostic approach has enabled us to support a variety of tool chains with minimal engineering work, but it does have some drawbacks. We implemented the project generation through an ad hoc mixture of makefile scripts and Python. This strategy makes the process difficult to debug, maintain, and extend. Our intent is for future versions to keep the concept of a master source-file list that only the makefile holds, but then delegate the actual generation to better-structured Python in a more maintainable way. 


%% file: tex/system_evaluation.tex
\section{System Evaluation} \label{sec:system_evaluation}

TFLM has undergone testing and it has been deployed extensively with many processors based on the Arm Cortex-M architecture \cite{armcortexm}. It has been ported to other architectures including ESP32 \cite{esp32} and many digital signal processors (DSPs). The framework is also available as an Arduino library. It can generate projects for environments such as Mbed~\cite{arm-mbed} as well. In this section, we use two representative platforms to assess and quantify TFLM’s computational and memory overheads. 

\subsection{Experimental Setup}

Our benchmarks focus on the (1) performance benefits of optimized kernels and (2) platforms we can support and the performance we achieve on them. So, we focus on extreme endpoints rather than on the overall spectrum. Specifically, we evaluate two extreme hardware designs and ML models.

We evaluate two extreme hardware designs: MCU (general) and ultra-low-power DSP (specialized). The details for the two hardware platforms are shown in Table~\ref{tab:eval_platforms}. First is the Sparkfun Edge, which has an Ambiq Apollo3 MCU. Apollo3 is powered by an Arm Cortex-M4 core and operates in burst mode at 96 MHz \cite{apollo3}. The second platform is an Xtensa Hifi Mini DSP, which is based on the Cadence Tensilica architecture \cite{tensilica}. 

\begin{table}[b]
\vspace{1em}
\centering
\resizebox{\columnwidth}{!}{
\begin{tabular}{|m{0.29\columnwidth}|m{0.22\columnwidth}|m{0.15\columnwidth}|m{0.15\columnwidth}|m{0.15\columnwidth}|}
\hline
\textbf{Platform}                                 & \textbf{Processor}    & \textbf{Clock} & \textbf{Flash}     & \textbf{RAM}       \\ \hline
Sparkfun Edge (Ambiq Apollo3) & Arm CPU \mbox{Cortex-M4}  & 96 MHz              & 1 MB       & 0.38 MB     \\ \hline
 Tensilica HiFi              & Xtensa DSP \mbox{HiFi Mini}  & 10 MHz        & 1 MB & 1 MB \\ \hline
\end{tabular}}
\caption{Embedded-platform benchmarking.}
\label{tab:eval_platforms}
\end{table}


We evaluate two extreme ML models in terms of model size and complexity for embedded devices. We use the Visual Wake Words (VWW) person-detection model \cite{chowdhery2019visual}, which represents a common microcontroller vision task of identifying whether a person appears in a given image. The model is trained and evaluated on images from the Microsoft COCO data set \cite{lin2014microsoft}. It primarily stresses and measures the performance of convolutional operations. Also, we use the Google Hotword model, which aids in detecting the key phrase ``OK Google.'' This model is designed to be small and fast enough to run constantly on a low-power DSP in smartphones and other devices with Google Assistant. Because it is proprietary, we use a version with scrambled weights and biases. More evaluation is better but TinyML is nascent and not many benchmarks exist. The benchmarks we use are part of TinyMLPerf~\cite{banbury2020benchmarking} and also used by MCUNet~\cite{lin2020mcunet}. 




Our benchmarks are INT8 TensorFlow Lite models in a serialized FlatBuffer format. The benchmarks run multiple inputs through a single model, measuring the time to process each input and produce an inference output. The benchmark does not measure the time necessary to bring up the model and configure the run time, since the recurring inference cost dominates total CPU cycles on most long-running systems.

\subsection{Benchmark Performance}


We provide two sets of benchmark results. First are the baseline results from running the benchmarks on reference kernels, which are simple operator-kernel implementations designed for readability rather than performance. Second are results for optimized kernels compared with the reference kernels. The optimized versions employ high-performance ARM CMSIS-NN and Cadence libraries \cite{lai2018cmsis}.

The results in Table~\ref{fig:perf_results_platforms} are for the CPU (Table~\ref{tab:inter_overhead_cpu}) and DSP (Table~\ref{tab:inter_overhead_dsp}). The total run time appears under the ``Total Cycles'' column, and the run time excluding the interpreter appears under the ``Calculation Cycles'' column. The difference between them is the minimal interpreter overhead.
The ``Interpreter Overhead'' column in both Table~\ref{tab:inter_overhead_cpu} and Table~\ref{tab:inter_overhead_dsp} is insignificant compared with the total model run time on both the CPU and DSP. The overhead on the microcontroller CPU (Table~\ref{tab:inter_overhead_cpu}) is less than 0.1\% for long-running models, such as VWW. In the case of short-running models such as Google Hotword, the overhead is still minimal at about 3\% to 4\%. The same general trend holds in Table~\ref{tab:inter_overhead_dsp} for non-CPU architectures like the Xtensa HiFi Mini DSP.

Comparing the reference kernel versions to the optimized kernel versions reveals considerable performance improvement. For example, between ``VWW Reference'' and ``VWW Optimized,'' the CMSIS-NN library offers more than a 4x speedup on the Cortex-M4 microcontroller. Optimization on the Xtensa HiFi Mini DSP offers a 7.7x speedup. For ``Google Hotword,'' the optimized kernel speed on Cortex-M4 is only 25\% better than the baseline reference model because less time goes to the kernel calculations. Each inner loop accounts for less time with respect to the total run time of the benchmark model. On the specialized DSP, the optimized kernels have a significant impact on performance. 

\begin{table*}[t]
    \begin{subtable}[t]{.5\textwidth}
        \centering
    \resizebox{\columnwidth}{!}{
    \begin{tabular}{|m{0.3\linewidth}|m{0.25\linewidth}|m{0.2\linewidth}|m{0.2\linewidth}|}
    \hline
    \textbf{Model}                    & \textbf{Total} \newline \textbf{Cycles} & \textbf{Calculation} \newline \textbf{Cycles}            & \textbf{Interpreter} \newline \textbf{Overhead} \\ \hline\hline
    VWW\newline Reference            & 18,990.8K    & 18,987.1K                     & \textless{}~0.1\%     \\ \hline
    VWW\newline Optimized            & 4,857.7K     & 4,852.9K                      & \textless{}~0.1\%     \\ \hline\hline
    Google Hotword Reference & 45.1K        & 43.7K                         & 3.3\%                \\ \hline
    Google Hotword Optimized & 36.4K        & 34.9K                         & 4.1\%                \\ \hline
    \end{tabular}}
    \caption{Sparkfun Edge (Apollo3 Cortex-M4)}
    \label{tab:inter_overhead_cpu}
    \end{subtable}%
    \hfill
   \begin{subtable}[t]{.5\textwidth}
        \centering
    \resizebox{\columnwidth}{!}{
    \begin{tabular}{|m{0.3\linewidth}|m{0.25\linewidth}|m{0.2\linewidth}|m{0.2\linewidth}|}
    \hline
    \textbf{Model}                    & \textbf{Total} \newline \textbf{Cycles} & \textbf{Calculation} \newline \textbf{Cycles}            & \textbf{Interpreter} \newline \textbf{Overhead} \\ \hline\hline
    VWW \newline Reference            & 387,341.8K   & 387,330.6K         & \textless{}~0.1\%     \\ \hline
    VWW \newline Optimized            & 49,952.3K    & 49,946.4K          & \textless{}~0.1\%     \\ \hline \hline
    Google Hotword Reference & 990.4K       & 987.4K             & 0.3\%                \\ \hline
    Google Hotword Optimized & 88.4K        & 84.6K              & 4.3\%                \\ \hline
    \end{tabular}}
    \caption{Xtensa HiFi Mini DSP}
    \label{tab:inter_overhead_dsp}
    \end{subtable}
    \caption{Performance results for TFLM target platforms.}
    \label{fig:perf_results_platforms}
\end{table*}

\subsection{Memory Overhead}

We assess TFLM's total memory usage. TFLM's memory usage includes the code size for the interpreter, memory allocator, memory planner, etc. plus any operators that are required by the model. Hence, the total memory usage varies greatly by the model. Large models and models with complex operators (e.g. VWW) consume more memory than their smaller counterparts like Google Hotword. In addition to VWW and Google Hotword, in this section, we added an even smaller reference convolution model containing just two convolution layers, a max-pooling layer, a dense layer, and an activation layer to emphasize the differences.

Overall, TFLM applications have a small footprint. The interpreter footprint, by itself, is less than 2KB (at max). 

Table~\ref{tab:model_size} shows that for the convolutional and Google Hotword models, the memory consumed is at most 13 KB. For the larger VWW model, the framework consumes 26.5 KB. 

To further analyze memory usage, recall that TFLM allocates program memory into two main sections: persistent and nonpersistent. Table~\ref{tab:model_size} reveals that depending on the model characteristics, one section can be larger than the other. The results show that we adjust to the needs of the different models while maintaining a small footprint.

\subsection{Benchmarking and Profiling}

TFLM provides a set of benchmarks and profiling APIs \cite{tfmicrobenchmarks} to compare hardware platforms and to let developers measure performance as well as identify opportunities for optimization. Benchmarks provide a consistent and fair way to measure hardware performance. MLPerf~\cite{reddi2020mlperf,mattson2020mlperf} adopted the TFLM benchmarks; the tinyMLPerf benchmark suite imposes accuracy metrics for them \cite{banbury2020benchmarking}.



Although benchmarks measure performance, profiling is necessary to gain useful insights into model behavior.
TFLM has hooks for developers to instrument specific code sections \cite{tfmicroprofiler}. 
These hooks allow a TinyML application developer to measure
overhead using a general-purpose interpreter rather than a custom neural-network engine for a specific model,
and they can examine a model's performance-critical paths.
These features allow identification, profiling, and optimization of bottleneck operators.
 
\begin{table}[t!]
\centering
\resizebox{\columnwidth}{!}{
\begin{tabular}{|m{0.3\columnwidth}|m{0.2\columnwidth}|m{0.27\columnwidth}|m{0.2\columnwidth}|}
\hline
\textbf{Model}                    & \textbf{Persistent}  \newline  \textbf{Memory} & \textbf{Nonpersistent} \newline  \textbf{Memory}                               & \textbf{Total} \newline \textbf{Memory} \\ \hline\hline
Convolutional \newline Reference           & 1.29 kB           & 7.75 kB & 9.04 kB      \\ \hline
Google Hotword \newline Reference & 12.12 kB           & 680 bytes                                                & 12.80 kB       \\ \hline
VWW \newline Reference &  26.50 kB          & 55.30 kB                                                & 81.79 kB        \\ \hline
\end{tabular}}
\caption{Memory consumption on Sparkfun Edge.}
\label{tab:model_size}
\end{table}

%% file: tex/related_work.tex
\section{Related Work} \label{sec:related_work}



There are a number of compiler frameworks for inference on TinyML systems. Examples include Microsoft's ELL~\cite{ms-ell}, which is a cross-compiler tool chain that enables users to run ML models on resource constrained platforms, similar to the platforms that we have evaluated. {Graph Lowering (GLOW)} \cite{rotem2018glow} is an open-source compiler that accelerates neural-network performance across a range of hardware platforms. {STM32Cube.AI} \cite{stm32cubeai} takes models from Keras, TensorFlow Lite, and others to generate code optimized for a range of STM32-series MCUs. {TinyEngine} \cite{lin2020mcunet} is a code-generator-based compiler that helps eliminate memory overhead for MCU deployments. {TVM} \cite{chen2018tvm} is an open-source ML compiler for CPUs, GPUs, and ML accelerators that has been ported to Cortex-M7 and other MCUs. {uTensor} \cite{utensor}, a precursor to TFLM, consists of an offline tool that translates a TensorFlow model into Arm microcontroller C++ machine code and it has a run time for execution management. 

In contrast to all of these related works, TFLM adopts a unique interpreter based approach for flexibility. An interpreter-based approach provides an alternative design point for others to consider when engineering their inference system to address the ecosystem challenges (Section~\ref{sec:technical_challenges}).

%% file: tex/conclusion.tex
\section{Conclusion} \label{sec:conclusion}

TFLM enables the transfer of deep learning onto embedded systems, significantly broadening the reach of ML. 
TFLM  is a framework that has been specifically engineered to run machine learning effectively and efficiently on embedded devices with only a few kilobytes of memory. TFLM's fundamental contributions are the design decisions that we made to address the unique challenges of embedded systems: hardware heterogeneity in the fragmented ecosystem, missing software features and severe resource constraints.




%% file: tex/acknowledgements.tex
\section*{Acknowledgements} \label{sec:acknowledgements}

TFLM is an open-source project and a community-based open-source project. As such, it rests on the work of many. We extend our gratitude to many individuals, teams, and organizations: Fredrik Knutsson and the CMSIS-NN team; Rod Crawford and Matthew Mattina from Arm; Raj Pawate from Cadence; Erich Plondke and Evgeni Gousef from Qualcomm; Jamie Campbell from Synopsys; Yair Siegel from Ceva; Sai Yelisetty from DSP Group; Zain Asgar from Stanford; Dan Situnayake from Edge Impulse; Neil Tan from the uTensor project; Sarah Sirajuddin, Rajat Monga, Jeff Dean, Andy Selle, Tim Davis, Megan Kacholia, Stella Laurenzo, Benoit Jacob, Dmitry Kalenichenko, Andrew Howard, Aakanksha Chowdhery, and Lawrence Chan from Google; and Radhika Ghosal, Sabrina Neuman, Mark Mazumder, and Colby Banbury from Harvard University.